\documentclass[letterpaper, 10 pt, conference]{ieeeconf}

\IEEEoverridecommandlockouts

\usepackage{graphicx}
\usepackage{acronym}
\usepackage{soul}
\usepackage{url}
\usepackage{hyperref}
\providecommand{\email}[1]{#1}

\acrodef{ros}[ROS]{Robot Operating System}
\acrodef{hri}[HRI]{human-robot interface}
\acrodef{bmi}[BMI]{brain-machine interface}
\acrodef{mpc}[MPC]{Model Predictive Control}
\acrodef{apf+vs}[ViSPF]{Virtual Sensors-Driven Potential Fields}
\acrodef{apf}[APFs]{Artificial Potential Fields}
\acrodef{at}[AT]{Assistive Technology}
\acrodef{vs}[VS]{Virtual Sensors}
\acrodef{vfh}[VFH]{Vector Field Histogram}
\acrodef{pq}[PQ]{Priority Queue}
\acrodef{b}[\textit{B}]{Baseline}
\acrodef{wo}[\textit{WO}]{Without OpenNav}
\acrodef{on}[\textit{ON}]{With OpenNav}
\acrodef{X}[\textit{D1}]{Desk 1}
\acrodef{Y}[\textit{D2}]{Desk 2}
\acrodef{Z}[\textit{P1}]{Person 1}
\acrodef{mlns}[MLNS]{Multi-Layer Navigation System}

\title{\LARGE \bf
    A Multi-Layer Navigation System for Smart Wheelchairs with a Limited Input Interface
}

\author{Piero Simonetto $^{1}$, Anna Polato$^{1}$, Francesco Pasti$^{1}$,Muhammad Rameez ur Rahman $^{2}$, Sebastiano Vascon $^{2,3}$,\\
Nicola Bellotto$^{1}$, Emanuele Menegatti$^{1, 4}$, Stefano Tortora$^{1}$, Luca Tonin$^{1,4}$%
\thanks{
This study was partially carried out within the PNRR research activities of the consortium iNEST funded by the European Union Next-GenerationEU (PNRR ECS\_00000043) and within the project PRINPNRR22 EasyWalk (Prot. P2022XJ7TE). This manuscript reflects only the authors’ views and opinions, neither the European Union nor the European Commission can be considered responsible for them.}%
\thanks{
$^{1}$Piero Simonetto, Anna Polato, Francesco Pasti, Stefano Tortora, Emanuele Menegatti, Luca Tonin are with Department of Information Engineering, University of Padova, Italy
\email{\{piero.simonetto, anna.polato.2, francesco.pasti \}@phd.unipd.it},
\email{\{stefano.tortora, nicola.bellotto, emanuele.menegatti, luca.tonin \}@unipd.it }%
}
\thanks{
$^{2}$ Muhammad Rameez ur Rahman $^{2}$, Sebastiano Vascon are part of the Ca' Foscari University of Venice
\email{\{muhammad.rahman,sebastiano.vascon\}@unive.it}
}
\thanks{
$^{3}$ Sebastiano Vascon are part of the European Centre for Living Technology
}
\thanks{
$^{4}$Emanuele Menegatti and Luca Tonin are also with the Padua Neuroscience Center, University of Padova, Italy}
}

\usepackage{xcolor}
\definecolor{ForestGreen}{RGB}{34, 139, 34}

\begin{document}

\maketitle
\thispagestyle{empty}
\pagestyle{empty}

\begin{abstract}
Smart wheelchairs are robotic devices that gather information from the surroundings and allow people to safely drive through various environments. They usually rely on the interpretation of high-level user's intentions, typically decoded through \ac{hri} or \ac{bmi}. However, navigating in an unknown environment and accurately understanding user commands remain significant challenges. Misinterpretation of commands or the environment can lead to user frustration, distrust of the system, or even unsafe situations. In this paper, we present a novel \ac{mlns} for a smart wheelchair that enables navigation in unknown environments while interpreting and safely contextualizing user's intentions even in the case of limited and faulty commands. The stack relies on a \ac{apf+vs} method for obstacle detection that accounts for the wheelchair's actual size, a \ac{mpc} module for trajectory generation, and an open-vocabulary computer vision algorithm for target identification. Finally, a system based on \ac{pq} was implemented for target selection. We validated our system on a real smart wheelchair with experiments involving 10 participants without disabilities, including scenarios with faulty input interfaces. The results demonstrate the robustness and effectiveness of our system in real-world environments, proving its potential for practical applications in assistive technologies.
\end{abstract}

\acresetall

\section{Introduction}
\ac{at} refers to any device and software designed to maintain and improve the functional capabilities of an individual, enabling their inclusion and participation in all areas of society. Recently, several studies have focused on the development of intelligent powered wheelchairs and on implementing appropriate \ac{hri} to enable people with severe motor or cognitive disabilities to control these systems~\cite{7947914,tonin2022learning, 7942067}.

Researchers have focused on hardware and software solutions that provide alternative interfaces to the traditional joystick and that allow end-users to communicate their intentions to the navigation system without using their upper-limbs. The most common approaches rely on deep learning techniques for detecting and interpreting user commands---e.g., hand gestures~\cite{10607210, sadi_finger-gesture_2022}, eye movement~\cite{xu_eye-gaze_2023}, and voice commands~\cite{iskanderani_voice_2021}. Other methods are based on the use of a joystick controlled by the lips~\cite{6733317}, tongue~\cite{mohammadi_high-resolution_2019}, or head-neck~\cite{zhang_using_2019}. Recently, similar commands have been generated by decoding brain activity using \ac{bmi}~\cite{tonin2022learning}.

\begin{figure}[t]
      \centering
          \hspace{-1.5cm}

        \includegraphics[width=0.47\textwidth]{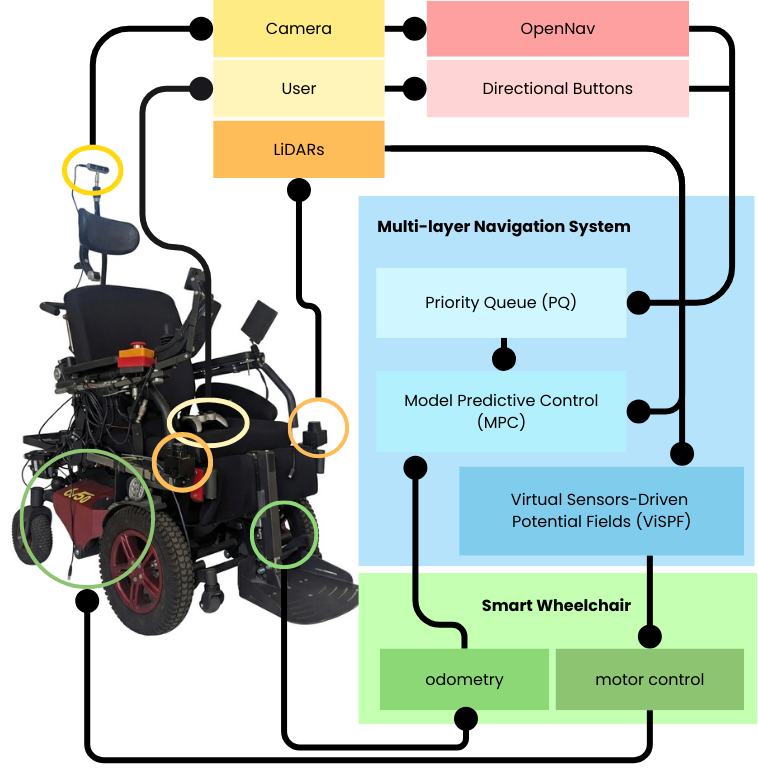}

      \caption{The proposed \acf{mlns} for a smart wheelchair semi-autonomous navigation. Given the user and OpenNav inputs, a \ac{pq} combined with \ac{mpc} and \ac{apf+vs} controls the navigation of the wheelchair.}
      \label{fig:sedia}
      \vspace{-0.5cm}
\end{figure}

Despite these advances, the integration of robotics in the assistive technology field is still at its infancy. Indeed, many solutions remain at the research or early validation stage, often demonstrating incremental improvements under controlled laboratory conditions. Their robustness and generalizability in real-world assistive scenarios are still limited. In practice, \ac{hri} for \ac{at} are highly sensitive to noise, environmental variability, sensor occlusions, and user-specific motor and cognitive fluctuations. As a consequence, they typically support discrete, high-level commands (e.g., left, right, stop, move) rather than continuous, intention-rich control signals. Although such abstractions simplify command decoding, they strongly constrain the interpretation of user intent. This limitation becomes particularly critical in navigation tasks involving close interaction with objects or people, where fine-grained adjustments, continuous modulation, and context awareness are required. In these situations, even small misclassifications or latency-induced errors may propagate into large trajectory deviations, reducing usability and increasing cognitive load. Therefore, understanding how to integrate such a limited input interface in the control of \ac{at} devices is not merely a performance optimization, but a prerequisite for effective navigation.

Furthermore, safety is a priority in these applications. Indeed, over the last few decades, many obstacle avoidance algorithms for smart wheelchairs have been evaluated. Among these, \ac{apf} have gained particular attention due to the ability to generate smooth trajectories and the straightforward implementation where attractive and repulsive forces are combined to guide the robot away from obstacles while steering it toward free spaces~\cite{borenstein1989real}.

In this work, we introduce a novel semi-autonomous navigation system for smart wheelchairs to operate in unknown environments with limited user input while maintaining adaptability and safety. The proposed architecture combines reactive obstacle avoidance, predictive trajectory planning, and computer vision algorithms for target identification within a framework designed for assistive mobility.

The remainder of the paper is structured as follows. Section~\ref{rworks} reviews related works, identifies the limitation of current approaches, and defines the aim of this research. Section~\ref{matmethods} presents the assistive platform and the developed navigation system, as well as the experimental protocol. Section~\ref{results} reports the results of the experiments followed by discussion, future work, and conclusions in Sections~\ref{discussion}-\ref{conclusion}.

\section{Related Works}\label{rworks}
As mentioned above, research in \ac{hri} for \ac{at} must address limitations that extend beyond technical performance. Repeated misinterpretations, unpredictable behaviors, or lack of transparency can compromise user trust in both the interface and the assistive system as a whole~\cite{firmino_de_souza_trust_2025}.
To mitigate the impact of noisy or unreliable user inputs, several works have shifted toward fully autonomous navigation paradigms~\cite{leishman2013driving,manta_apf-based_2021}.
In these systems, the user typically specifies only a high-level destination, while the robot autonomously manages local motion planning and obstacle avoidance. Although this strategy reduces cognitive load and increases robustness against command misinterpretation, it also reduces direct user involvement in the navigation process, potentially affecting perceived agency and acceptance.
Empirical evidence suggests that many users prefer semi-autonomous or shared-control strategies, where assistance enhances safety without removing control authority~\cite{yang_preserving_2025}.
Within this framework, several approaches rely on adaptive assistance mechanisms in which the level of autonomy is modulated according to the user’s needs or explicit requests~\cite{schettino2020improving,devigne_shared_2019}.

In these systems, machine intervention is triggered in the presence of risk, instability, or degraded performance.
Some shared-control architectures move beyond event-based arbitration and instead maintain a continuous contribution from both the user and the autonomous system. Rather than alternating authority, these approaches blend human intent and autonomous intelligence in real time, establishing a cooperative interaction~\cite{devigne_low_2016, nguyen_shared_2013}.
A critical requirement in such frameworks is the reliable estimation of user intention. In this direction, Casado et al. propose an intention-prediction method based on a generative diffusion model for short-term trajectory planning in a shared-control smart wheelchair paradigm~\cite{casado_navigating_2026}. The technique estimates user intention by jointly modeling user inputs and environmental information, producing motion plans that reflect both contributions.

Similarly, several works propose an alternative shared-control strategy based on Bayesian inference~\cite{deng_bayesian_2020, trieu_shared_2008, trieu_adaptive_2009}. More advanced solutions work directly with neural signals to decode user's intentions by exploiting \ac{bmi}-driven shared control to drive smart wheelchairs or telepresence robots~\cite{tonin2022learning,7109829,LEEB2013121}. Their probabilistic formulation explicitly accounts for uncertainty in intention decoding, reinforcing the central role of intention prediction in cooperative navigation.

In the context of assistive technologies and smart wheelchairs, classical reactive collision avoidance strategies remain widely adopted. The most investigated approaches are the \ac{vfh}~\cite{bell_assistive_1994}, the Dynamic Window Approach (DWA)~\cite{goos_navchair_1994,adamek_enhancing_2022}, and \ac{apf}.
While \ac{vfh} and DWA provide structured velocity selection mechanisms, \ac{apf} remain particularly attractive in assistive robotics due to their intuitive formulation, smooth motion generation, and low computational cost. However, \ac{apf} methods are also affected by well-known limitations, extensively analyzed in foundational works such as~\cite{koren_potential_1991}. These include local minima, where attractive and repulsive forces balance out, oscillatory behavior in narrow corridors or doorways, sensitivity to parameter tuning, and susceptibility to sensor noise.
Several assistive robotics works have attempted to adapt and enhance \ac{apf} formulations to overcome these limitations.
For example, in \cite{tonin2022learning} authors implemented a \ac{apf}-based navigation strategy in a \ac{bmi}-driven smart wheelchair and validated it with end-users. Although the system demonstrated feasibility in controlled environments, oscillations in narrow passages emerged due to sensor noise and the intrinsic limitations of the \ac{apf} formulation.
Other works have adapted \ac{apf}-based navigation to shared-control frameworks, demonstrating their feasibility in hands-free interfaces and reporting improvements in safety and user experience metrics, including reduced perceived cognitive workload~\cite{chen_enabling_2025,ashley_semi-autonomous_2017}.
A more modern direction involves integrating predictive control mechanisms into the navigation stack. \ac{mpc} has been adopted as a trajectory planner in smart wheelchairs~\cite{ceravolo2017model}.

\ac{mpc} is particularly suitable for tracking trajectories that resemble natural user motion patterns~\cite{de2024nonlinear}, while explicitly handling system constraints. Still, its deployment in assistive platforms presents several structural limitations. The wheelchair is usually modeled as a unicycle robot, which reduces computational complexity and facilitates real-time implementation, but it neglects relevant dynamic effects. As a result, prediction accuracy may be degraded, potentially affecting trajectory optimality and tracking performance. As a consequence, collision avoidance is encoded in the distance between the obstacle and the center of the robot, ignoring its real geometry. This may create incoherence between the full range of feasible paths and the set computed by the collision avoidance system. Finally, the computational time strictly depends on the spatial configuration, density, and geometry of the obstacles in the environment. In cluttered scenarios or constrained passages, a higher number of active constraints can significantly increase the complexity of the optimization problem and slow down convergence, compromising real-time execution.

The problem of accurately representing the wheelchair's physical footprint has been addressed through simplified geometric hardware-based approximations. Most approaches model the platform as either circular or elliptical~\cite{chatzidimitriadis_evaluation_2017, 9901691} to facilitate analytical treatment and reduce computational complexity. Alternatively, some systems rely on multiple proximity sensors distributed along the perimeter of the wheelchair (e.g., ultrasonic arrays) to approximate its occupied space during navigation \cite{ashley_semi-autonomous_2017,diao_design_2017}.
One of the earliest works explicitly addressing wheelchair geometry in navigation is \cite{levine_navchair_1999}, which proposed modeling the platform through a discrete set of representative points and introduced the use of five control points for computing repulsive forces within a \ac{vfh}-based framework.

Motivated by the limitations discussed above, this work addresses three main gaps in current assistive navigation systems.
First, we introduce an enhanced \ac{apf+vs} formulation that explicitly accounts for the actual geometry and footprint of the wheelchair, overcoming the simplifications associated with center-based collision constraints and circular approximations, without compromising real-time performance. Second, we propose a multi-layer navigation architecture that integrates \ac{apf+vs} and \ac{mpc} to leverage their complementary strengths while addressing the computational burden typically associated with MPC-based navigation, improving real-time feasibility even in cluttered environments. Third, we develop an \ac{hri} framework based on a \ac{pq} mechanism and a computer vision module to improve intention inference and environmental awareness. Finally, we experimentally evaluated the robustness of the proposed approach under noisy input conditions characteristic of multiple common \acp{hri} for \ac{at}.
The system was implemented and tested on a real smart wheelchair platform in semi-autonomous assistive scenarios.

\section{Materials and Methods}\label{matmethods}

\subsection{Customized powered wheelchair}
In this work, we customized a commercially available powered wheelchair (18.68E HI-LO Electric, Vassilli s.r.l., Padua, Italy) as shown in Figure~\ref{fig:sedia}. The velocity and position of the wheelchair were tracked using two Kübler KIS40 encoders attached to the front wheels by belt and the resulting odometry was computed according  to~\cite{xu2019path}. Two Hokuyo URG-04LX-UG01 LIDARs were mounted in front of the wheelchair to detect obstacles, while a RealSense D435i camera, placed above the user's head, captured the environment with a field of view similar to the user's perspective. The user controlled the wheelchair using a Logitech F710 joystick. All components were connected to a Lenovo Thinkpad T15g Gen 2, 11th Gen Intel Core i9-11950H @ 2.60 GHz x 16, 32 GB RAM, NVIDIA GeForce RTX 3080. We used ROS Noetic 1.16 to provide a common interface for custom components. An R-Net chipset (Curtiss-Wright Corporation, Davidson, United States) was used to connect the R-Net CAN-bus network of the wheelchair to the external laptop.

\subsection{\acf{mlns}}

The architecture of \ac{mlns} is reported in Figure~\ref{fig:sedia}. It is composed of three main elements: a custom \ac{apf+vs} layer, to ensure proactive and safe navigation in real-time, a \ac{mpc} layer used to build smooth trajectories, and a \ac{pq} layer to fuse user's commands and decisions autonomously taken by the intelligent navigation system. The \ac{pq} relies on visual information provided by OpenNav~\cite{rahman2024opennavefficientopenvocabulary}, a computer vision module composed of YOLO-World~\cite{cheng2024yolow} and MobileSAM~\cite{zhao2023fast} and designed for real-time open vocabulary 3D object detection.
The navigation stack computes the velocity command in a real-time reconstructed local map that tracks the possible obstacles. The map was used by both the \ac{mpc} and \ac{apf+vs} control layers. Since the wheelchair occludes its backside to the lasers, a spatio-temporal voxel plugin~\cite{vsplros} was used to track obstacles in this area. By construction, the field of view of the laser in this setup did not ensure that the space right behind the wheelchair was empty. For this reason, we decided to inhibit direct reverse commands.

\subsection{\acf{apf+vs} Layer}\label{sub:vispf}
\begin{figure}[t]
      \centering
       \includegraphics[scale=0.2]{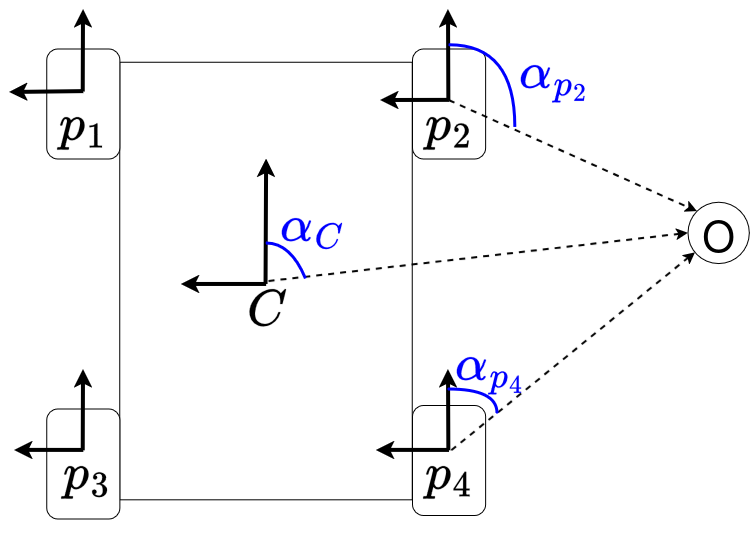}
      \caption{Graphical representation of the proposed \ac{apf+vs}. The wheelchair is represented with a rectangle surrounded by its four wheels. C represents the center of the wheelchair and \(p_{\{1,2,3,4\}}\) the new \ac{vs}.\(
      \alpha_{\{C,p_1,p_2,p_3,p_4\}}\).}
      \label{fig:apf}
      \vspace{-0.5cm}
\end{figure}

To ensure a suitable clearance between the obstacles and the robot’s entire footprint, and to prevent collisions with any part of it, we modified the \ac{apf} layer to reproduce some \textit{\ac{vs}}. In this new method, reported in Figure~\ref{fig:apf}, instead of modeling the wheelchair as a unique point, each vertex of the rectangle area of the wheelchair \(p_i\), \(i \in \{1,2,3,4\} \) is subject to a linear and angular velocity based on its position with respect to the nearest obstacles (repellers) around the wheelchair, with an approach similar to that reported in~\cite{tonin2022learning}. During navigation, the obstacle positions are provided in polar coordinates in \(p_i \) centered reference frame. In order to set an upper limit on the computation time, the angular values were discretized into \(N=36\)
sectors.
Linear velocity is calculated as
\begin{equation}
    v_{l,i}(t) = V_{TMAX}(1-e^{-\frac{d_{O,i}-D_F}{\tau }})
\end{equation}
where \(V_{TMAX}\) is the maximum velocity (0.2~m/s), \(D_F \) is the safety distance. \(d_{O,i}\) is the distance detected between the nearest obstacle and \(p_i\) in its range of motion and \(\tau\) is a smoothing parameter. The resulting linear velocity is  the minimum among the four calculated velocities.
At the same time, the angular velocity is calculated as
\begin{equation}
    v_{\alpha,i}(t) = \sum_{k=1}^N \lambda_k (\alpha_{p_i}-\alpha_C)e^{-\frac{(\alpha_{p_i}-\alpha_C)^2}{2\sigma_k^2}}
\end{equation}
where \(N\) is the number of angular sectors and \(\lambda_k\) depends on the distance (\(d_k\)) of the nearest obstacle in sector $k$ from the center of the wheelchair:
\begin{equation}
    \lambda_k = e^{-\frac{d_k-D_F}{\tau}}
\end{equation}

and \(\alpha_{p,i}\) and \(\alpha_C\) are the angles of the obstacle with respect to the point \(p_i\) and the center of the robot \(C\). \(\sigma_k\) represents the strength of the detected repeller. Recalling that \(\alpha=0\) implies moving in a straight direction, the subtraction \((\alpha_{p_i}-\alpha_C)\) defines the sign of the velocity value.

The final angular velocity is computed as the sum of all components:
\begin{equation}
    v_\alpha(t) = \sum_{i=1}^4 v_{\alpha,i}(t) + v_a
\end{equation}
with \(v_a\) the attractor's angular component.

This method allows navigation to adapt to the wheelchair's shape and dimension and ensures precision in collision avoidance.

\begin{figure}[th]
      \centering
        \includegraphics[scale=0.3]{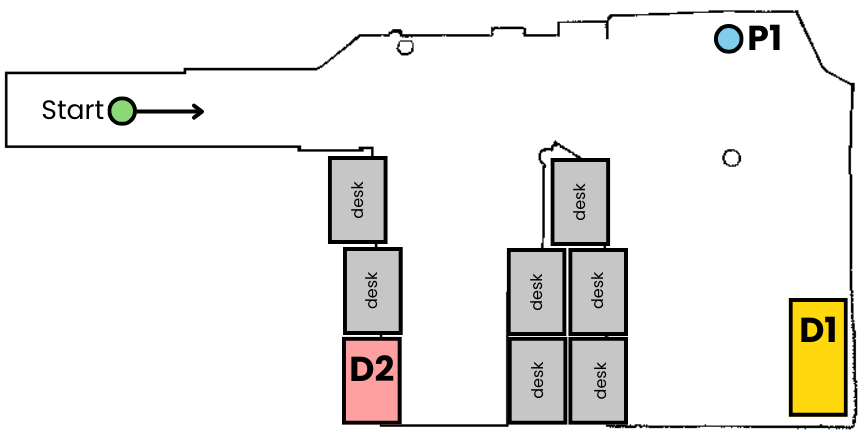}
      \caption{Map of the laboratory where the experiment took place. The green point is the starting position, Target \ac{X} and \ac{Y} are desks, and Target \ac{Z} is a person.}
      \label{fig:map}
      \vspace{-0.5cm}
\end{figure}

\subsection{\ac{mpc} and trajectory generation}
A \ac{mpc} is exploited to generate a trajectory from the current wheelchair position to a given target (see next section). Our implementation of the \ac{mpc} is based on the problem defined in Eq. 3 of~\cite{de2024nonlinear}. In this formulation, obstacles are identified by clustering the wheelchair's surroundings and representing them as discrete points at a distance that accounts for the wheelchair's physical dimensions. To ensure that the solver generates reliable solutions, specific constraints have been implemented: the maximum number of obstacles is limited to 50, and the planning horizon is restricted to the next 30 seconds of motion.

Conversely to traditional approaches---where the \ac{mpc} is used to directly provide the command velocity---the generated path is sub-sampled at fixed intervals of 0.3~m and the resulting samples are considered as attractors for the \ac{apf+vs} layer. This solution ensures a safe control even if environmental dynamics change faster and prevents the classical oscillations of the artificial potential fields.

\subsection{Target selection: \acf{pq} \label{sub:goal_selection}}

\ac{mpc} requires a target to model a path.
The choice of target is expected from the \ac{pq} that periodically selects the current goal among the available ones. We provide the following two implementations. \ac{pq} \textit{A} exposes two targets: the one sent by the user and the one sent by default by the system. The user's input is the first choice and overcomes the default behavior. \ac{pq} \textit{B} exposes three targets: The user's one, the default's one, and a target provided by the computer vision module OpenNav. The OpenNav target stands in the middle, it overcomes the default behavior but follows the user's input.

Regarding the type of target, the user sends it via a limited input interface based on the joystick described below. OpenNav sets a target on a surrounding object, the closest to the moving direction of the wheelchair, as described above. Eventually, the default target point is set in front of the wheelchair, 2~meters ahead, every 4~seconds.

In this work, we experimentally evaluated both \acp{pq}, investigating how OpenNav can influence their usability and navigation itself in the presence of misinterpreted commands.

\subsection{Limited input interface}
The \ac{mlns} is tested on experiments in which the user communicates to the robot through a limited input interface. More precisely, the user can choose only between two turning directions (i.e., left and right) as in the case of head-neck joystick. The direction is selected through the directional buttons of a joystick and transformed into a point in the space around the wheelchair. Furthermore, we introduced  a minimum refractory period of 3~seconds between consecutive user commands to emulate anti-tremors filters usually implemented in this kind of interfaces. Finally we designed a perturbation process that reverts the user's command with a fixed probability of $0.2$ to simulate wrong button presses.

\subsection{Experiments  \label{experiments}}
The semi-autonomous navigation system of the smart wheelchair was evaluated by 10 users without disabilities (3 females, 7 males, 27 ± 3 years old). Users sat on the wheelchair and were asked to navigate from a fixed starting point to a target destination (identified by a common object) by sending commands to the wheelchair using the joystick directional buttons.

Each user was asked to navigate from a starting point A to one of the targets reported in Figure~\ref{fig:map} (i.e., two desks and one person identified respectively by letters \ac{X}, \ac{Y}, \ac{Z} on the map). The users autonomously decided when the target was reached by stopping the navigation.

Each user performed 12 navigation runs for each target, 4 for each of the three following modalities based on \ac{pq} policies (Section~\ref{sub:goal_selection}):
\begin{itemize}
    \item \ac{b}: user's input priority (\ac{pq}~\textit{A}), no perturbations;
    \item \ac{wo}: user's input priority (\ac{pq}~\textit{A}), with perturbations;
    \item \ac{on}: user's input or OpenNav priority (\ac{pq}~\textit{B}), with perturbations.
\end{itemize}
In the first modality~\ac{b}, users had full control of the wheelchair and no errors were introduced in the input interface. Conversely, in \ac{wo} and \ac{on} users evaluated the system with a faulty input interface and without and with the OpenNav module, respectively.
During navigation, participants were blinded to the active condition, which was randomized across trials.
In \ac{on}, OpenNav was configured to detect the "desk" for Target \ac{X} and \ac{Y} and the "person" for Target \ac{Z}.

\section{Results}~\label{results}

First, it is worth to mention that the \ac{mlns} allowed the wheelchair to move in a cluttered real environment without any crashes and among 360 trials conducted by 10 participants, the system achieved a 99.72\% task completion rate, with only one interrupted run (Target \ac{Y}, approach \ac{on}) after user request.

\begin{figure*}[th!]
      \centering
        \includegraphics[width=0.89\textwidth]{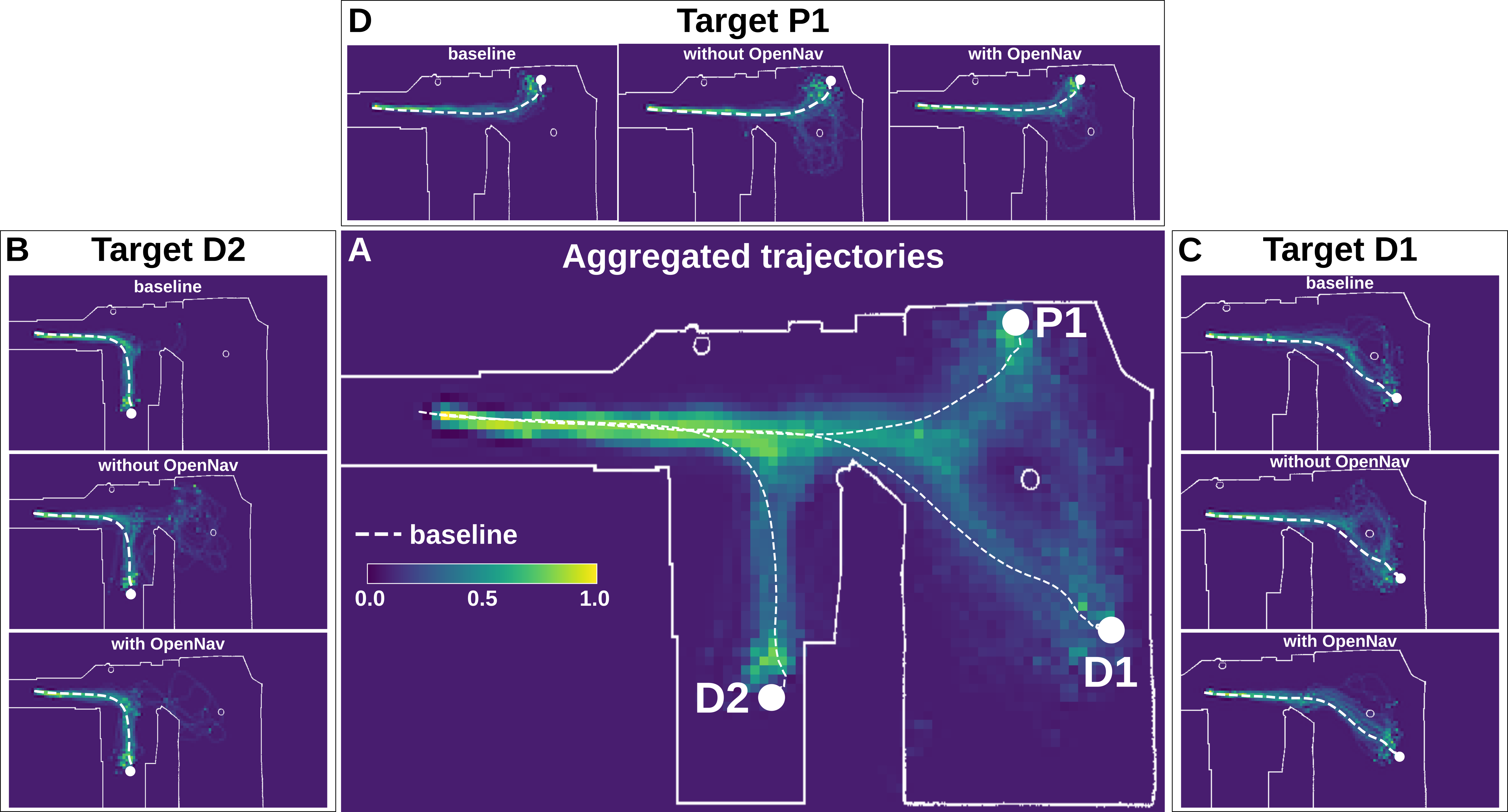}

      \caption{A) Normalized heatmaps of the aggregated trajectories produced during the experiments (center). B-D) Normalized heatmaps of the trajectories divided per target and modality.
      White dashed lines represent the average Baseline path for each target.
      Heatmap represents a color gradient of the crossed positions,
      from the more (in yellow) to the less (in blue) common.}
      \label{fig:tr_full}
\end{figure*}

Figure~\ref{fig:tr_full}A presents the normalized heatmap aggregating all user trajectories (central panel). The white dashed lines represent the average \ac{b} trajectories for each target. The spatial distribution highlights the dominant motion paths emerging across trials, providing a global view of path consistency and dispersion.
Figures~\ref{fig:tr_full}B-D illustrate the results separately for each target (\ac{X}, \ac{Y}, and \ac{Z}) and for each modality (\ac{b}, \ac{wo}, and \ac{on}). Heatmaps reveal clear differences in spatial dispersion. \ac{wo} trajectories exhibit greater variability, with occasional deviations far from the nominal path. These paths correspond to runs in which perturbations in user commands led to unintended deviations. In contrast, the implementations of \ac{b} and \ac{on} result in a visibly reduced dispersion and a more concentrated flow of trajectories toward the target. This suggests that perception-driven assistance effectively constrains the set of feasible motion strategies, promoting more repeatable and stable navigation behavior, similar to the baseline (i.e., without input perturbations).

\begin{figure*}[thpb]
      \centering
        \includegraphics[width=\textwidth]{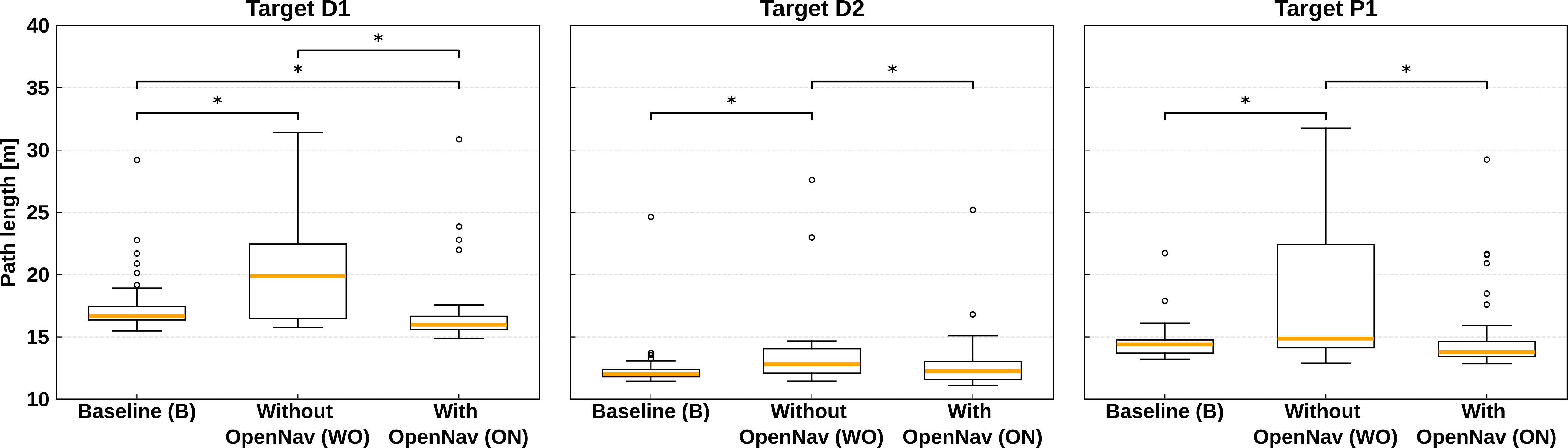}
      \caption{Distribution of the paths length traveled for each target and navigation modality. Horizontal brackets indicate pairs with a statistically significant difference (Kruskal-Wallis with Dunn's post-hoc test, $p < 0.05$)}
      \label{fig:path_length}
\end{figure*}

To further quantify the efficiency of the system and consistency of the generated paths, Figure~\ref{fig:path_length} depicts the distribution of all paths length for each target and navigation modality. For target \ac{X}, the modality \ac{on} shows a statistically relevant decrease ($\sim0.6m$, $p<0.05$, Bonferroni corrected) in the paths length compared to \ac{b}. Furthermore, across all targets, the modality \ac{wo} displays significantly larger ranges and more pronounced outliers compared to both other methods ($p<0.05$, Bonferroni corrected). These results indicate that while the \ac{b} and \ac{on} pipelines maintain a steady and predictable motion path, the \ac{wo} configuration is prone to occasional substantial deviations and less efficient trajectories, as confirmed by the higher standard deviations reported in Table~\ref{tab:distance_mean}.

\begin{figure*}[thpb]
      \centering
        \includegraphics[width=\textwidth]{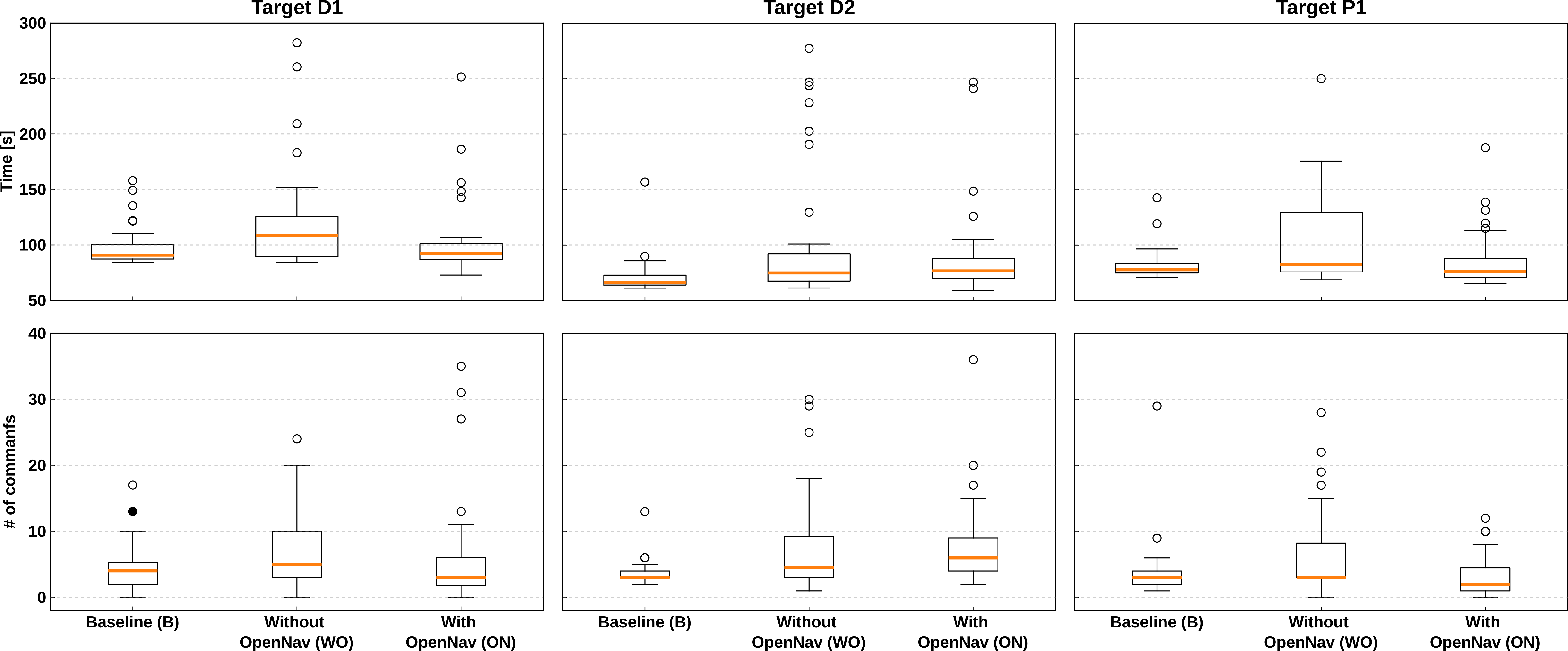}

      \caption{Distribution of the completion time (top) and number of commands sent by the user (bottom) of each target.}
      \label{fig:tr_sep}
    \vspace{-0.4cm}

\end{figure*}

Figure~\ref{fig:tr_sep} presents the distributions of completion time (top) and the number of user commands (bottom) for the three methods (\ac{b}, \ac{wo}, \ac{on}). In targets \ac{X} and \ac{Z}, \ac{on} exhibits a lower mean completion time and a narrower inter-quartile range compared to \ac{wo}, indicating not only faster execution, but also reduced variability between runs. This suggests that perception-driven assistance promotes more consistent performance between users and trials. A similar trend is observed in the distribution of user commands: \ac{on} generally reduces the number of corrective inputs required, reflecting a lower interaction burden and a smoother control process. For target \ac{Y}, the differences between the two assisted conditions are less pronounced. The mean completion times are comparable between \ac{on} and \ac{wo} modalities (77.3 vs. 74.0~s, Table~\ref{tab:mean_time}), and the average number of commands is slightly higher in the \ac{on} modality (5.8 vs 4.7, Table~\ref{tab:mean_cmd}).

\begin{table}[t]
\caption{Mean and standard deviation of completion time per target and method}
\vspace{-0.5cm}
\label{tab:mean_time}
\begin{center}
\begin{tabular}{c|c|c|c}
       & \ac{b} & \ac{wo} & \ac{on} \\
       \hline
 \ac{X} & 92.1 \(\pm\) 17.3   & 108.4  \(\pm\) 44.1   & 91.5  \(\pm\) 31.5       \\
\hline
 \ac{Y} & 69.9  \(\pm\) 15.5   & 74.0  \(\pm\) 65.2     & 77.3 \(\pm\) 38.3        \\
\hline
 \ac{Z} & 77.7 \(\pm\) 13.8    & 81.7  \(\pm\) 39.2     & 77.7  \(\pm\) 23.9  \\
\hline
\end{tabular}
\end{center}
\vspace{-0.2cm}

\end{table}

\begin{table}[t]
\caption{Mean and standard deviation of number of commands per target and method}
\vspace{-0.5cm}
\label{tab:mean_cmd}
\begin{center}
\begin{tabular}{c|c|c|c}
       & \ac{b} & \ac{wo} & \ac{on} \\
\hline
 \ac{X} & 4  \(\pm\) 3.5   & 5.3 \(\pm\) 5.4     & 3.1  \(\pm\) 7.6     \\
\hline
 \ac{Y} & 3.2  \(\pm\) 1.8   & 4.7 \(\pm\) 9.4      & 5.8  \(\pm\) 6.1      \\
\hline
 \ac{Z} & 3.1 \(\pm\) 4.5    & 3.9  \(\pm\) 6.1     & 2   \(\pm\) 2.9    \\
\hline
\end{tabular}
\end{center}
\vspace{-0.8cm}
\end{table}

\begin{table}[t]
\caption{Mean and standard deviation of length of the paths per target and method}
\vspace{-0.5cm}

\label{tab:distance_mean}
\begin{center}
\begin{tabular}{c|c|c|c}
       & \ac{b} & \ac{wo} & \ac{on} \\
\hline
 \ac{X} & 16.6766 \(\pm\) 2.5074 & 19.8891  \(\pm\)  7.1318   & 15.9831  \(\pm\)  4.2086     \\
\hline
 \ac{Y} & 12.0068  \(\pm\)  2.0267 & 12.7959  \(\pm\)  10.3488  & 12.2596  \(\pm\)  5.7333      \\
\hline
 \ac{Z} & 14.3877  \(\pm\)  1.5002   &  14.8688  \(\pm\)  6.6185   & 13.7696  \(\pm\)  3.2384   \\
\hline
\end{tabular}
\end{center}
\vspace{-0.8cm}
\end{table}

\section{Discussion}~\label{discussion}

The proposed \ac{mlns} showed robust performance in an unknown environment, effectively mitigating the well-known limitations of both \ac{mpc} and \ac{apf} approaches. With the \ac{apf+vs} method that provides reactive obstacle avoidance and \ac{mpc} that allows predictive trajectory optimization, their integration enables the system to take advantage of the complementary strengths of both algorithms. The execution of user commands relies on the \ac{mpc} process, which introduces computation-dependent latency. During this short planning interval, the wheelchair maintains the last \ac{mpc} calculated trajectory while the \ac{apf+vs} layer continues to enforce local collision avoidance constraints, with a processing rate of 50~Hz. This layered interaction ensures continuous safety and ensures a stable and smooth navigation experience. In addition, the cooperation between the predictive \ac{mpc} layer and the proposed reactive \ac{apf+vs} layer overcomes the classical oscillatory behaviors typically observed with pure \ac{apf}-based navigation, especially in confined environments such as narrow corridors or door passages. Furthermore, the introduction of \ac{vs} (Section~\ref{sub:vispf}) enables the experimental trials to be conducted without any collisions while ensuring that the entire footprint of the smart wheelchair maintains a consistent safe clearance from the surrounding objects and walls.
A video of the above-mentioned navigation experience is available in the paper attachments.

The integration of OpenNav into \ac{pq} proved to be an effective design choice. Across all tasks, the trajectories generated with \ac{on} remain closer to the \ac{b} than those obtained with \ac{wo}, indicating stronger alignment with the user-driven paths of \ac{b} (not affected by perturbations). In general, the pipeline \ac{wo} exhibits greater spatial dispersion, reflecting increased variability in navigation behavior. For targets \ac{X} and \ac{Z}, the efficiency gains introduced by OpenNav are evident. Compared to the corresponding configuration \ac{wo}, users require fewer corrective commands and achieve shorter completion times, while maintaining trajectory profiles that closely resemble the baseline. This suggests that OpenNav reduces the interaction effort without compromising user intent. Nevertheless, target \ac{Y} requires additional discussions. The task consists of approaching a table located at the end of a corridor, with additional tables positioned along both sides. Two critical phases characterize this scenario: the initial turn required to enter the corridor and the selection of the target table. The corridor entry maneuver is strongly dependent on the initial user command in combination with the collision avoidance system. When this first input is inaccurate, due to the limited input interface, users may deviate significantly from the nominal path and require substantial corrective actions to realign with the corridor entrance. This behavior is consistent with the outliers observed in Figure~\ref{fig:tr_full}, as both \ac{on} and \ac{wo} configurations, affected by perturbations, exhibit some trajectories that diverge markedly from the baseline. Regarding table selection, the presence of multiple nearby tables may lead OpenNav to prioritize the closest feasible target.
In such cases, users may issue additional commands to override or refine the system’s choice and continue toward the intended table. This interaction dynamic highlights a balanced human-robot synergy. Within this framework, the human component becomes essential in resolving environmental ambiguities. This dynamic explains the slightly higher number of commands and longer execution times observed for target \ac{Y} under the OpenNav condition (Section~\ref{results}), and is a clear reflection of the shared-control paradigm adopted in this study and of the collaborative nature of the system.

Overall, OpenNav is particularly beneficial during the early and mid phases of navigation, when the user is still far from the target and can assess and, if necessary, refine the system’s choice. Its benefit decreases during the final approach to the target locations, when users typically prefer to manually adjust the wheelchair’s orientation for precise alignment.

\section{Conclusions \label{conclusion}}
In this paper, we present and experimentally validate a novel multi-layer navigation framework for smart wheelchairs that integrates a new \acl{apf+vs} layer, \acl{mpc}, and OpenNav computer vision algorithm within a unified architecture. The proposed system enables the smart wheelchair to plan motions in space accounting for the entire footprint of the wheelchair. Moreover, we introduce and evaluate two priority queue-based pipelines to regulate a hierarchy of control commands while preserving user authority at the highest level. The \acp{pq} are designed to evaluate the system in the case of a limited input interface. We tested the setup with a real device and with 10 users in order to verify the performance. Experimental results indicate that users were able to navigate the environment successfully, while relying on the assistive system, suggesting effective human–robot cooperation under constrained control conditions. The user maintains control of the wheelchair, with the only exception of a not reachable direction. In this case, the system tends to carry the user to the nearest safe position. The configuration incorporating OpenNav demonstrated clear advantages in tasks involving distant targets, enabling simpler and more direct trajectories, shorter completion times, and requiring fewer corrective inputs.
Future studies will include validation on a larger population of users and in more complex and heterogeneous environments, in order to evaluate the system performance under realistic assistive conditions and to further assess its robustness and adaptability.

\bibliographystyle{IEEEtran}
\bibliography{IEEEabrv,bibiograpy/bibiograpy.bib}

\end{document}